\def\year{2021}\relax
\documentclass[letterpaper]{article} 
\usepackage{aaai21}  
\usepackage{times}  
\usepackage{helvet} 
\usepackage{courier}  
\usepackage[hyphens]{url}  
\usepackage{graphicx} 
\usepackage{natbib}  
\usepackage{caption} 
\usepackage{amsmath, amssymb}
\usepackage{multirow}
\usepackage{array}
\usepackage{mwe}
\usepackage{graphbox}
\usepackage[printonlyused,withpage]{acronym}
\newacro{gan}[GAN]{Generative Adversarial Network}
\newacro{ae}[AE]{Autoencoders}
\newacro{vae}[VAE]{Variational Autoencoder}
\newacro{dae}[DAE]{Denoising Autoencoder}
\newacro{pgd}[PGD]{Projected Gradient Descent}
\newacro{sgd}[SGD]{Stochastic Gradient Descent}
\newacro{relu}[ReLU]{Rectified Linear Unit}
\newacro{mse}[MSE]{Mean Squared Error}
\newacro{arae}[ARAE]{Adversarially Robust trained Autoencoder}

\DeclareMathOperator{\E}{\mathbb{E}}
\DeclareMathOperator*{\argmax}{arg\,max}

\title{ARAE: Adversarially Robust Training of Autoencoders Improves \\Novelty Detection }
\author{
Mohammadreza Salehi,\textsuperscript{\rm 1} 
Atrin Arya,\textsuperscript{\rm 1} 
Barbod Pajoum,\textsuperscript{\rm 1} 
Mohammad Otoofi,\textsuperscript{\rm 1}\\ 
Amirreza Shaeiri,\textsuperscript{\rm 1} 
Mohammad Hossein Rohban,\textsuperscript{\rm 1} 
Hamid R. Rabiee\textsuperscript{\rm 1}\\
}
\affiliations{

    \textsuperscript{\rm 1}Computer Engineering Department, Sharif University of Technology, Tehran, Iran. \\
{smrsalehi, aarya, pajoum}@ce.sharif.edu,
m.otoofi@ictic.sharif.edu,\\
shairi@ce.sharif.edu,
rohban@sharif.edu,
rabiee@sharif.edu
}

\begin{document}

\maketitle

\begin{abstract}
\ac{ae} have recently been widely employed to approach the novelty detection problem. Trained only on the normal data, the AE is expected to reconstruct the normal data effectively while failing to regenerate the anomalous data. Based on this assumption, one could utilize the AE for novelty detection. However, it is known that this assumption does not always hold. More specifically, such an AE can often perfectly reconstruct the anomalous data as well, due to modeling of low-level and generic features in the input. To address this problem, we propose a novel training algorithm for the AE that facilitates learning of more semantically meaningful features. For this purpose, we exploit the fact that adversarial robustness promotes learning of meaningful features. Therefore, we force the AE to learn such features by making its bottleneck layer more stable against adversarial perturbations. This idea is general and can be applied to other autoencoder based approaches as well. We show that despite using a much simpler architecture in comparison to the prior methods, the proposed AE outperforms or is competitive to state-of-the-art on four benchmark datasets and two medical datasets.
\end{abstract}

\section{Introduction}

\begin{table*}[t]
\centering
\begin{tabular}{c}
\includegraphics[width = 14cm]{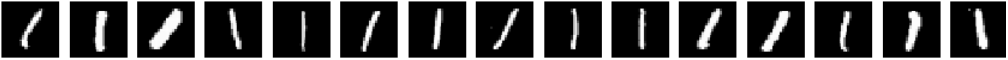} \\
\includegraphics[width = 14cm]{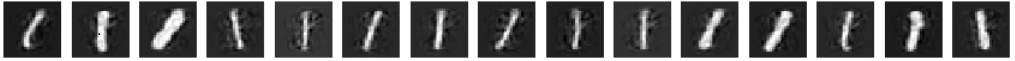}\\
\includegraphics[width = 14cm]{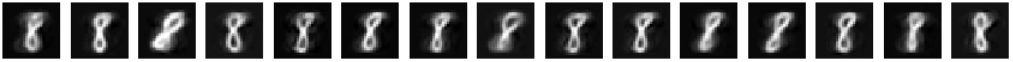}
\end{tabular}
\captionof{figure}{Unlike \ac{dae}, \ac{arae} that is trained on the normal class, which is the digit $8$, reconstructs a normal instance when it is given an anomalous digit, from the class $1$. The first row shows the input images. The second and third rows show the \ac{dae} and \ac{arae} reconstructions of the corresponding inputs, respectively. \ac{arae} is trained based on bounded $\ell_\infty$, $\ell_2$, rotation, and translation perturbations.} \label{fig:output}
\end{table*}
\raggedbottom

In many real-world problems, it is easy to gather normal data from the operating behavior of a system. However, collecting data from the same system in situations where it malfunctions or is being used clumsily may be difficult or even impossible. For instance, in a surveillance camera that captures daily activity in an environment, almost all frames are related to the normal behavior. This means that data associated with the anomalous behavior is difficult to obtain from such cameras. Anomaly/novelty detection refers to the set of solutions for such settings.

The key point in the definition of anomaly detection is the outlier notion. In the literature, An outlier is defined as a data point that deviates from the bulk of the remaining data \cite{hawkins1980identification, chalapathy2019deep}. Assuming that the normal data is generated by a distribution, the goal is to detect whether a new unseen observation is drawn from this distribution or not. 
In prior work, \ac{ae} and \ac{gan} were extensively applied for novelty detection \cite{sabokrou2018adversarially, perera2019ocgan, schlegl2017unsupervised, akcay2018ganomaly}. 

In \ac{gan}-based approaches, one tries to train a model that could adversarially generate realistic images from the normal class. This means that if the model fails to generate a given input image, the input would probably be an anomalous one. However, \ac{gan}-based approaches face some challenges during the training. These include mode collapse that happens when the generator maps several inputs to a single image in the output space. In \ac{gan}, complete mode collapse is rare, while a partial collapse occurs more frequently \cite{goodfellow2016nips, kodali2017convergence}. Furthermore, high sensitivity of the training to the choices of hyperparameters, non-convergence problem, parameter oscillation, and non-reproducible results due to the unstable training are counted as the other challenges in training of the \ac{gan} \cite{martin2017towards, salimans2016improved}. 

On the other hand, \ac{ae} is more convenient to train and gives results that are easier to reproduce. Therefore, we propose our method based on \ac{ae}-based approaches in this paper. An \ac{ae}, which has learned features that are mostly unique to the normal class, could reconstruct the normal data perfectly, while when given an anomalous data, it either reconstructs a corrupted or a normal output; In the former case, the anomalous input is likely to have disjoint features compared to the normal class, while in the latter, the input may resemble a normal data in some aspects. Note that in both cases, unlike for the normal data, the reconstruction \ac{mse}  is high for the anomalous data. This means that for such an \ac{ae}, we could threshold the reconstruction loss to distinguish the normal vs. anomalous data. One could alternatively leverage a discriminator that is applied to the reconstructed image to distinguish between the anomalous and normal data \cite{sabokrou2018adversarially, larsen2015autoencoding}. In any case, as mentioned, an important premise for the \ac{ae} to work is that it learns mostly unique features to the normal class. We call such features ``semantically meaningful" or ``robust", contrasted with generic low level features that are subject to change in presence of noise, in the rest of the paper.

A common problem in using \ac{ae} for novelty detection is its generalization ability to reconstruct some anomaly inputs, when they share common features with the normal class \cite{gong2019memorizing, zong2018deep}. 
Although this generalization property is useful in other contexts, such as restoration \cite{mao2016image}, it is considered as a drawback in novelty detection. In other papers \cite{hasan2016learning, zhao2017spatio, sultani2018real}, the main underlying assumption behind the \ac{ae}-based approaches is that the reconstruction error is high when the model is given an anomalous data, which as mentioned does not seem to be holding perfectly. 

There are two reasons why the main underlying assumption in these methods does not hold necessarily. First, the model behavior when facing the anomalous data is not observed and is not therefore predictable. Second, the learned latent space may capture mostly the features that are in common between the normal and anomalous data. When given the anomalous data, this would likely yield a perfectly reconstructed anomalous data. 
To address these issues, we aimed for a solution that learns an adversarially robust latent space, where the focus is on learning unique or semantically meaningful features of the normal inputs and their nuances. This could prevent the decoder from reconstructing the anomalies. 

It is shown in \cite{madry2017towards} that small imperceptible changes in the input can easily fool a deep neural network classifier. \ac{ae}'s are subject to such attacks as well. This stems from the fact that a deep classifier or an \ac{ae} would likely learn low level or brittle non-robust features \cite{ilyas2019adversarial}. Low level features could be exploited to reconstruct {\it any} given image perfectly. Hence, the presence of such features seems to violate the main underlying assumption of the earlier work for novelty detection that is based on \ac{ae}. Therefore, we propose to train an adversarially robust \ac{ae} to overcome this issue. In Figure \ref{fig:output}, reconstructions from \ac{dae} and the proposed method are shown. Here, the normal data is considered to be the number $8$ in the MNIST dataset and the models are trained only on the normal category. As opposed to the proposed \ac{arae}, \ac{dae} generalizes and reconstructs the number $1$ perfectly. This is not desired in the novelty detection problem. This means that the latent space of \ac{dae} has learned features that are not necessarily meaningful. 

To train a robust \ac{ae} for the novelty detection task, a new objective function based on adversarial attacks is proposed. The novel \ac{ae} which is based on a simple architecture, is evaluated on MNIST, Fashion-MNIST, COIL-100, CIFAR-10, and two medical datasets. We will next review existing approaches in more details, and then describe our proposed idea along with its evaluation. We demonstrate that despite the simplicity of the underlying model, the proposed model outperforms or stays competitive with state-of-the-art in novelty detection. Moreover, we show that our method performs much better compared to another state-of-the-art method in presence of adversarial examples, which is more suitable for real-world applications. 

\section{Related work}

\begin{figure*}[t]
  \centering
  \includegraphics[width=0.7\linewidth]{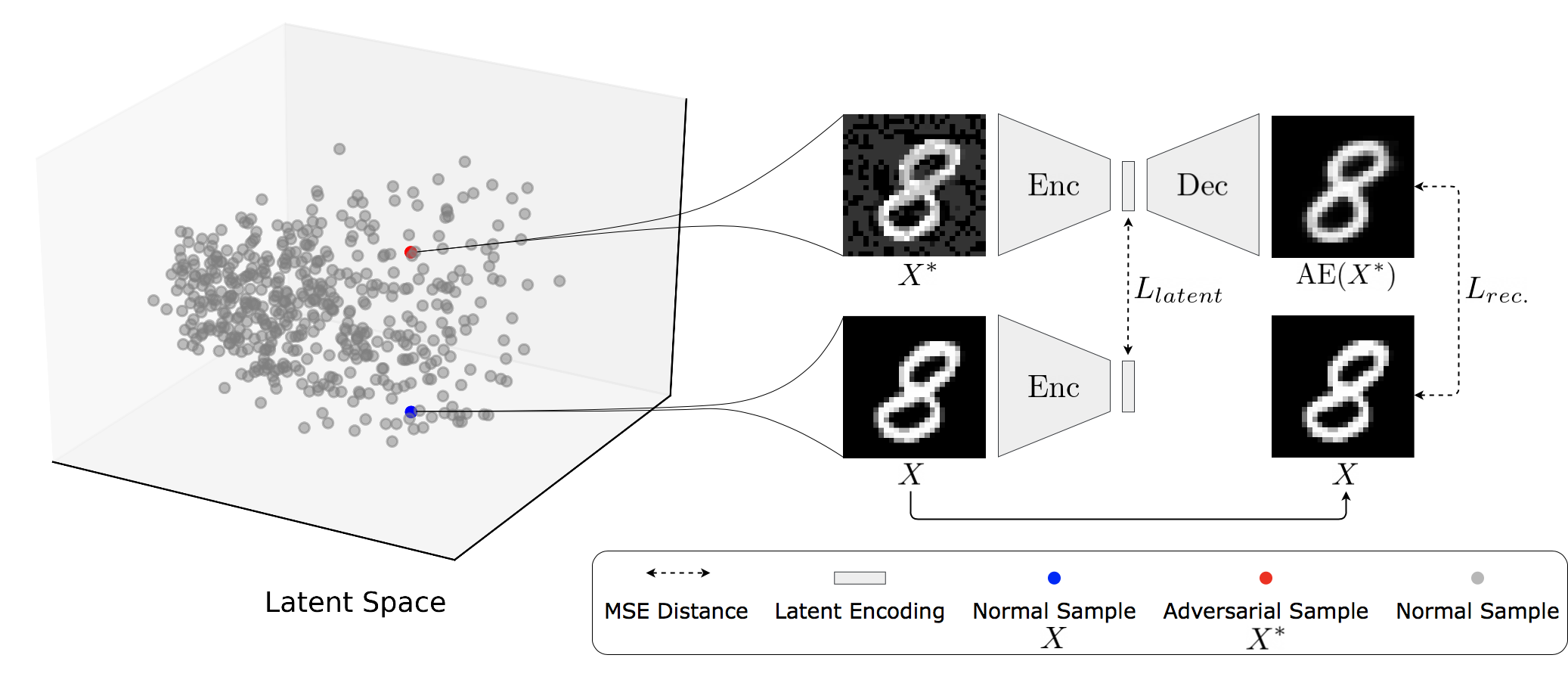}
  \caption{The training procedure of our method. $L_{latent}$ and $L_{rec.}$ are obtained using the \ac{mse} distance and used to form $L_{AE}$.}
  \label{fig:train}
\end{figure*}

As explained earlier in the introduction, methods that are used in the literature are classified into two main categories: (1) modeling the normal behavior in the latent space; and (2) thresholding the \ac{ae} reconstruction error. Of course, a hybrid of these two approaches was also considered in the field.

DRAE \cite{zhou2017anomaly}, takes the second approach, i.e. it is based on the \ac{mse} distance between the \ac{ae} output and its input. An underlying assumption in this work is that the training data may contain abnormal samples. Therefore, the method tries to identify these samples throughout the training process. It finally uses only the reconstruction error in the test time.

As an extension to the \ac{ae}-based methods, in OCGAN \cite{perera2019ocgan}, a model is introduced in which the \ac{ae} is trained by using 4 \ac{gan}s, a classifier, and the ``negative sample mining" technique. Here, both the encoder and decoder of the \ac{ae} are considered as generators in the \ac{gan}. At the inference time, the method only uses \ac{mse} between the model output and input to make a prediction. The authors attempted to force the encoder output distribution to be approximately uniform. They also forced the decoder output distribution to resemble the normal input distribution in the whole latent domain. This is expected to result in a higher \ac{mse} distance between the decoder output and input for the abnormal data. This method achieved state-of-the-art results at the time of presentation.

\cite{abati2019latent} and \cite{sabokrou2018adversarially} are the other examples in the \ac{ae}-based approaches, except that in \cite{abati2019latent}, additionally, the probability distribution over the latent space was obtained for the normal input data. Then, in the test time, the probability of a sample being normal, which is called the ``surprise score", is added to the reconstruction error before the thresholding happens. 
In \cite{sabokrou2018adversarially},  there is a possibility of using the discriminator output, which is a real number between zero and one, as an alternative to the \ac{mse} distance in order to find the anomaly score. This is done by considering the \ac{ae} as the generator in the \ac{gan} framework.

In \cite{pidhorskyi2018generative}, a \ac{gan} is initially used to obtain the latent space, then the probability distribution of the normal class over the latent space is considered to be as the multiplication of two marginal distributions, which are learned empirically. \cite{ruff2018deep} (DSVDD) tries to model the normal latent space with the presumption that all normal data can be compressed into a hyper-sphere. This framework can be considered as a combination of Deep Learning and classical models such as \cite{chen2001one} (One-class SVM), that has the advantage of extracting more relevant features from the training data than the above-mentioned \cite{chen2001one} because the whole network training process is done in an end-to-end procedure. In \cite{schlegl2017unsupervised}, a \ac{gan} framework is used to model the latent space. It is assumed that if the test data is normal, then a sample could be found in a latent space such that the corresponding image that is made by the generator is classified as real by the \ac{gan} discriminator. 

\section{Method}

\begin{figure*}[t]
  \centering
  \includegraphics[width=0.75\linewidth]{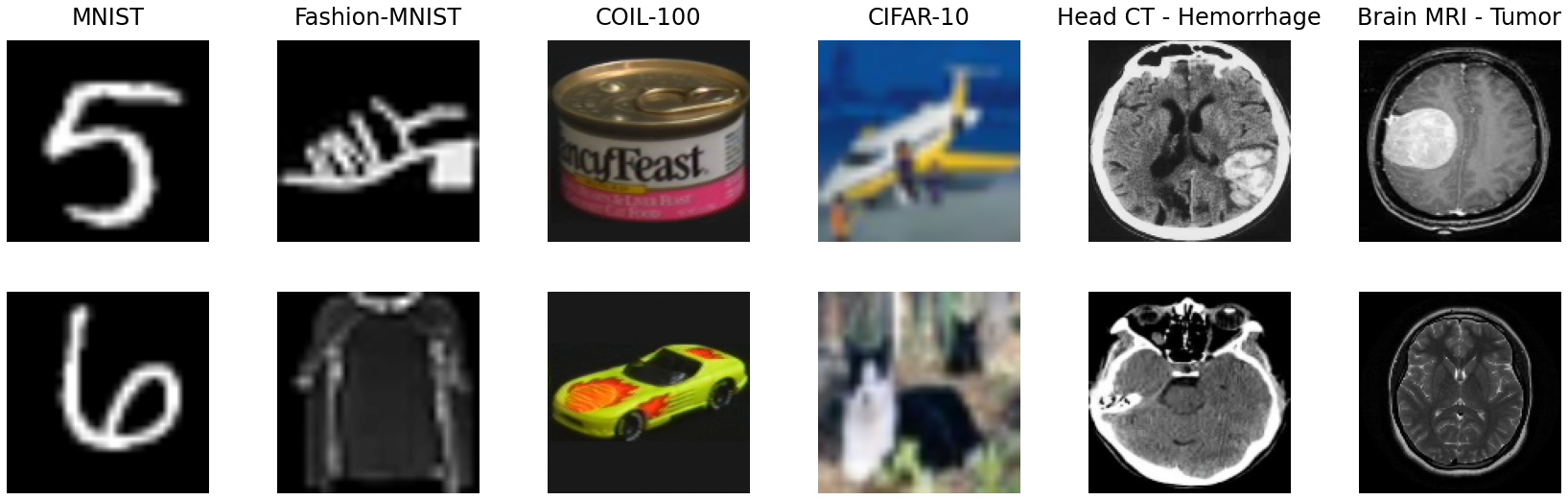}
  \caption{Samples from the evaluation datasets. For the medical datasets, the top row samples are anomalous and the bottom row samples are normal.}
  \label{fig:datasets}
\end{figure*}

As we discussed earlier, the main problem of \ac{ae} is its strong generalization ability. We observe that \ac{dae} does not necessarily learn distinctive features of the normal class. To remedy this problem, our approach is to force the \ac{ae} latent space implicitly to model only unique features of the normal class. To make this happen, the framework for adversarial robustness, which is proposed in \cite{madry2017towards, ilyas2019adversarial}, is adopted. We propose to successively craft adversarial examples and then utilize them to train the \ac{ae}. Adversarial examples are considered as those irrelevant small changes in the input that destabilize the latent encoding. We will next describe the details of the proposed adversarial training in the following sections. The training procedure is demonstrated in \hbox{Figure \ref{fig:train}.}

\subsection{Adversarial Examples Crafting}
In a semantically meaningful latent space, two highly perceptually similar samples should share similar feature encodings. Therefore, searching for a sample $X^*$ that is perceptually similar to a sample $X$, but has a distant latent encoding from that of $X$, leads us to an adversarial sample. As opposed to the normal sample $X$, the adversarial sample $X^*$ is very likely to have a high reconstruction loss, thus it would be detected as abnormal by the \ac{ae}, despite being perceptually similar to a normal sample. Therefore, based on this intuition, the following method is used to craft the adversarial samples.

At the training epoch $i$, we craft a set of adversarial samples $S^i_{(adv)}$ based on the initial training dataset $S$. For this purpose, we slightly perturb each sample $X \in S$ to craft an adversarial sample $X^*$ that has two properties: (1) $X^*$ is perceptually similar to $X$, through controlling the $\ell_\infty$ distance of $X$ and $X^*$; (2) $X^*$ latent encoding is as far as possible from that of $X$. This is equivalent to solving the following optimization problem:
\begin{equation} 
\max_{\delta_X}L_{\text{latent}}\mbox{ s.t. }{\|\delta_X\|}_{\infty} \leq \epsilon
\end{equation}
\begin{equation} \label{adv_ex_cr}
L_{\text{latent}} = \|\mbox{Enc}(X+\delta_X)-\mbox{Enc}(X)\|^2_2\
\end{equation}
In this formulation, ${\|\ .\ \|}_p$ is the $\ell_p$-norm, $\epsilon$ is the attack magnitude, and $X^* = X + \delta_X$ is the adversarial sample. We solve this optimization problem for each sample $X\in S$ using the  \ac{pgd} \cite{madry2017towards} method, to obtain $S^i_{(adv)}$. 

\subsection{Autoencoder Adversarial Training}
To train the \ac{ae} using the crafted dataset $S^i_{(adv)}$ in the previous section, we propose the following loss function:
\begin{equation}
L_{\text{AE}} = L_{\text{rec.}} + \gamma L_{\text{latent}}
\end{equation}
where $\gamma$ is a balancing hyperparameter, $L_{\text{latent}}$ refers to the loss function that is introduced in Eq. \ref{adv_ex_cr} and $L_{\text{rec.}}$ corresponds to the following loss function:
\begin{equation}
L_{\text{rec.}} = \|X - \mbox{Dec}(\mbox{Enc}(X^*))\|^2_2\
\end{equation}

At each step, the \ac{ae} is trained one epoch on the adversarially crafted samples using this loss function.
In the training procedure, the $L_{\text{rec.}}$ term forces the \ac{ae} to reconstruct the adversarial samples properly, while the $L_{\text{latent}}$ term forces the adversarial samples to have closer representations to that of the corresponding normal samples in the latent space. We observe that the encoder decreases $L_{\text{latent}}$ to a limited extent by merely encoding the whole input space into a compact latent space. Too compact latent space results in a high $L_{\text{rec.}}$, which is not achievable when the network is trained using $L_{\text{AE}}$. A compact latent space causes the latent encodings of anomalous data to be close to that of normal data. Thus for any given input, the generated image is more likely to be a normal sample. To summarize, the whole training procedure is trying to solve the following saddle point problem \cite{wald1945statistical}:
\begin{equation} \label{summ_eq}
\begin{gathered}
 \delta^*_X := \argmax_{\|\delta_X\|_\infty \le \epsilon} L_{\text{latent}}(X, \delta_X, W) \\
 \min_{W}\E_{X}\left[\gamma L_{\text{latent}}(X, \delta^*_X, W) + L_{\text{rec.}}(X, \delta^*_X, W)\right]
\end{gathered}
\end{equation}
where $W$ is denoted as the \ac{ae} weights. Note that it was shown that the adversarial training could not be solved in a single shot by the \ac{sgd}, and one instead should try other optimization algorithms such as the \ac{pgd}. This relies on Danskin theorem to solve the inner optimization followed by the outer optimization \cite{madry2017towards}.

\section{Experiments}

In this section, we evaluate our method, which is denoted by \ac{arae}, and compare it with state-of-the-art on common benchmark datasets that are used for the unsupervised novelty detection task. Moreover, we use two medical datasets to evaluate our method in real-world settings. We show that even though our method is based on a simple and efficient architecture, it performs competitively or superior compared to state-of-the-art approaches. Furthermore, we provide insights about the robustness of our method against adversarial attacks. The results are based on several evaluation strategies that are used in the literature. All results that are reported in this paper are reproducible by our publicly available implementation in the Keras framework \cite{chollet2015}\footnote{\url{https://github.com/rohban-lab/Salehi_submitted_2020}}. 

\subsection{Experimental Setup}
\subsubsection{Baselines}
Baseline and state-of-the-art approaches like VAE \cite{kingma2013auto}, OCSVM \cite{chen2001one}, AnoGAN \cite{schlegl2017unsupervised}, DSVDD \cite{ruff2018deep}, MTQM \cite{wang2019multivariate}, OCGAN \cite{perera2019ocgan}, LSA \cite{abati2019latent}, DAGMM \cite{zong2018deep}, DSEBM \cite{zhai2016deep}, GPND \cite{pidhorskyi2018generative}, $l_1$ thresholding \cite{soltanolkotabi2012geometric}, DPCP \cite{tsakiris2018dual}, OutlierPursuit \cite{xu2010robust}, ALOCC \cite{sabokrou2018adversarially}, LOF \cite{breunig2000lof}, and DRAE \cite{xia2015learning} are selected to be compared with our method. Results of some of these methods were obtained from \cite{perera2019ocgan, wang2019multivariate, pidhorskyi2018generative}.
\subsubsection{Datasets}
We evaluate our method on MNIST \cite{lecun2010mnist}, Fashion-MNIST \cite{xiao2017fashion}, COIL-100 \cite{Nene96objectimage}, CIFAR-10 \cite{Krizhevsky2009cifar}, Head CT - hemorrhage \cite{kitamura2018hemorrhage}, and Brain MRI - Tumor \cite{chakrabarty2019tumor} datasets. Samples from each dataset are shown in Figure \ref{fig:datasets}. These datasets differ in size, image shape, complexity and diversity. Next, we briefly introduce each of these datasets.
\begin{itemize}
    \item MNIST: This dataset contains 70,000 $28\times28$ grayscale handwritten digits from 0 to 9.
    \item Fashion-MNIST: A dataset similar to MNIST with 70,000 $28\times28$ grayscale images of 10 fashion product categories.
    \item CIFAR-10: This dataset contains 60000 $32\times32$ color images of 10 categories.
    \item COIL-100: A dataset of 7200 color images of 100 different object classes. Each class contains 72 images of one object captured in different poses. We downscale the images of this dataset to the size $32\times 32$.
    \item Head CT - Hemorrhage: A dataset with 100 normal head CT slices and 100 other with 4 different kinds of hemorrhage. Each slice comes from a different person and the image size is $128\times 128$. 
    \item Brain MRI - Tumor: A dataset with 253 brain MRI images. 155 of them contain brain tumors and the rest 98 are normal. The image size is $256\times 256$.
\end{itemize}

\subsubsection{Protocols}

To carry out the training-testing procedure, we need to define the data partitions. For MNIST, Fashion-MNIST, and CIFAR-10, one class is considered as the normal class and samples from the other classes are assumed to be anomalous. For COIL-100, we randomly take $n$ classes as the normal classes, where $n\in \{ 1,4,7 \}$, and use the samples from the remaining classes as the anomalous samples. For the mentioned dataset, this process is repeated 30 times and the results are averaged. For the medical datasets, the brain images with no damage are considered as the normal class and the rest form the anomalous class. To form the training and testing data, there are two protocols that are commonly used in the framework of unsupervised novelty detection\cite{pidhorskyi2018generative, perera2019ocgan, sabokrou2018adversarially}, which are as follows:
\begin{itemize}
    \item Protocol 1: The original training-testing splits of the dataset are merged, shuffled, and $80\%$ of the normal class samples are used to train the model. The remaining $20\%$ forms some specified portion (denoted as $\tau$) of the testing data. The other portion is formed by randomly sampling from the anomalous classes.
    \item Protocol 2: The original training-testing splits of the dataset are used to train and test the model. The training is carried out using the normal samples and the entire testing data is used for evaluation.
\end{itemize}
We compare our method to other approaches using Area Under the Curve (AUC) of the Receiver Operating Characteristics (ROC) curve, the $F_1$ score, and the False Positive Rate (FPR) at $99.5\%$ True Positive Rate (TPR). Here, we let the positive class be the anomalous one unless otherwise specified. 
\subsubsection{Architecture and Hyperparameters}

Our \ac{ae} uses a 3-layer fully connected network with layer sizes of $(512, 256, 128)$, following the input-layer to encode the input. A decoder, whose architecture is mirroring that of the encoder, is used to reconstruct the output. Each layer of the network is followed by a sigmoid activation. This architecture is used for all the datasets except the medical ones and CIFAR-10. For the medical datasets and CIFAR-10, we use a convolutional \ac{ae} which is explained in \cite{bergman2019improving}. For datasets with complex and detailed images like COIL-100, Fashion-MNIST, CIFAR-10, and the medical datasets, the hyperparameter $\epsilon$, which is the maximum perturbation $\ell_\infty$ norm as defined in Eq. \ref{summ_eq}, is set to $0.05$, while for MNIST it is set to $0.2$. The hyperparameter $\gamma$, defined in Eq. \ref{summ_eq}, is always set to $0.1$.

\subsection{Results}

\begin{table}[t]
\centering
\caption{AUC values (in percentage) for the medical datasets. The standard deviation of the last 50 epochs' AUCs are included for the Brain MRI - Tumor dataset.}
\label{table:medical}
{\small
\begin{tabular}{cccccc}
\hline\noalign{\smallskip}
{Dataset} & {OCGAN} & {LSA} & {ARAE}\\
\hline
\noalign{\smallskip}
Head CT - Hemorrhage & 51.2 & 81.6  & \textbf{84.8} \\
\noalign{\smallskip}
\multirow{2}{*}{Brain MRI - Tumor} & 91.7 & 95.6 & \textbf{97.0} \\
& $\pm 3$ & $\pm 1.4$ & $\pm 0.5$ \\
\hline
\end{tabular}}
\end{table}
\raggedbottom

\begin{table*}[t]
\centering
\caption{AUC values (in percentage) on MNIST and FMNIST (Fashion-MNIST). The standard deviation of the last 50 epochs' AUCs are included for our method on MNIST. The values were obtained for each class using protocol 2.}
\label{table:mnist}
{\small
\begin{tabular*}{\textwidth}{c @{\extracolsep{\fill}} c @{\extracolsep{\fill}} c @{\extracolsep{\fill}} c @{\extracolsep{\fill}} c @{\extracolsep{\fill}} c @{\extracolsep{\fill}} c @{\extracolsep{\fill}} c @{\extracolsep{\fill}} c @{\extracolsep{\fill}} c @{\extracolsep{\fill}} c @{\extracolsep{\fill}} c @{\extracolsep{\fill}} c}
\hline\noalign{\smallskip}
Dataset & Method & 0 & 1 & 2 & 3 & 4 & 5 & 6 & 7 & 8 & 9 & Mean\\
\hline
\noalign{\smallskip}
\multirow{9}{*}{MNIST} & \multicolumn{1}{|c}{VAE} & 98.5 & 99.7 & 94.3 & 91.6 & 94.5 & 92.9 & 97.7 & 97.5 & 86.4 & 96.7 & 95.0\\
& \multicolumn{1}{|c}{OCSVM} & 99.5 & 99.9 & 92.6 & 93.6 & 96.7 & 95.5 & 98.7 & 96.6 & 90.3 & 96.2 & 96.0\\
& \multicolumn{1}{|c}{AnoGAN} & 96.6 & 99.2 & 85.0 & 88.7 & 89.4 & 88.3 & 94.7 & 93.5 & 84.9 & 92.4 & 91.3\\
& \multicolumn{1}{|c}{DSVDD} & 98.0 & 99.7 & 91.7 & 91.9 & 94.9 & 88.5 & 98.3 & 94.6 & 93.9 & 96.5 & 94.8\\
& \multicolumn{1}{|c}{MTQM} & 99.5 & 99.8 & 95.3 & 96.3 & 96.6 & 96.2 & 99.2 & 96.9 & 95.5 & 97.7 & 97.3\\
& \multicolumn{1}{|c}{OCGAN} & 99.8 & 99.9 & 94.2 & 96.3 & 97.5 & 98.0 & 99.1 & 98.1 & 93.9 & 98.1 & \textbf{97.5}\\
& \multicolumn{1}{|c}{LSA} & 99.3 & 99.9 & 95.9 & 96.6 & 95.6 & 96.4 & 99.4 & 98.0 & 95.3 & 98.1 & \textbf{97.5}\\
\noalign{\smallskip}
\cline{2-13}
\noalign{\smallskip}
 & \multicolumn{1}{|c}{\multirow{2}{*}{ARAE}} & 99.8 & 99.9 & 96.0 & 97.2 & 97.0 & 97.4 & 99.5 & 96.9 & 92.4 & 98.5 & \textbf{97.5}\\
 & \multicolumn{1}{|c}{} & $\pm 0.017$ & $\pm 0.003$ & $\pm 0.2$ & $\pm 0.17$ & $\pm 0.14$ & $\pm 0.1$ & $\pm 0.03$ & $\pm 0.1$ & $\pm 0.3$ & $\pm 0.04$ & $\pm 0.04$\\
 \noalign{\smallskip}
\hline
\noalign{\smallskip}
\multirow{7}{*}{FMNIST} & \multicolumn{1}{|c}{VAE} & 87.4 & 97.7 & 81.6 & 91.2 & 87.2 & 91.6 & 73.8 & 97.6 & 79.5 & 96.5 & 88.4\\
&  \multicolumn{1}{|c}{OCSVM} & 91.9 & 99.0 & 89.4 & 94.2 & 90.7 & 91.8 & 83.4 & 98.8 & 90.3 & 98.2 & 92.8\\
&  \multicolumn{1}{|c}{DAGMM} & 30.3 & 31.1 & 47.5 & 48.1 & 49.9 & 41.3 & 42.0 & 37.4 & 51.8 & 37.8 & 41.7\\
&  \multicolumn{1}{|c}{DSEBM} & 89.1 & 56.0 & 86.1 & 90.3 & 88.4 & 85.9 & 78.2 & 98.1 & 86.5 & 96.7 & 85.5\\
&  \multicolumn{1}{|c}{MTQM} & 92.2 & 95.8 & 89.9 & 93.0 & 92.2 & 89.4 & 84.4 & 98.0 & 94.5 & 98.3 & 92.8\\
&  \multicolumn{1}{|c}{LSA} & 91.6 & 98.3 & 87.8 & 92.3 & 89.7 & 90.7 & 84.1 & 97.7 & 91.0 & 98.4 & 92.2\\
\noalign{\smallskip}
\cline{2-13}
\noalign{\smallskip}
&  \multicolumn{1}{|c}{ARAE} & 93.7 & 99.1 & 91.1 & 94.4 & 92.3 & 91.4 & 83.6 & 98.9 & 93.9 & 97.9 & \textbf{93.6}\\
\noalign{\smallskip}
\hline
\end{tabular*}}
\end{table*}
\raggedbottom

\begin{table*}[t]
\centering
\caption{AUC and $F_1$ values on the COIL-100 dataset. The values were obtained using protocol 1 for $n\in \{1,4,7\}$ and different $\tau$s, where n and $\tau$ represent the number of normal classes and the testing data portion of the normal samples, respectively.}
\label{table:coil}
{\small
\begin{tabular*}{\textwidth}{c @{\extracolsep{\fill}} c @{\extracolsep{\fill}} c @{\extracolsep{\fill}} c @{\extracolsep{\fill}} c @{\extracolsep{\fill}} c @{\extracolsep{\fill}} c}
\hline\noalign{\smallskip}
Parameters & Metric & { OutlierPursuit} & {DPCP} & { $l_1$ thresholding} & { GPND} & { ARAE}\\
\hline
\noalign{\smallskip}
\multicolumn{1}{c|}{\multirow{2}{*}{$n=1$, $\tau=50\%$}} & AUC & 0.908 & 0.900 & 0.991 & 0.968 & \textbf{0.998}\\
\multicolumn{1}{c|}{} & $F_1$ & 0.902 & 0.882 & 0.978 & 0.979 & \textbf{0.993}\\
\noalign{\smallskip}
\hline
\noalign{\smallskip}
\multicolumn{1}{c|}{\multirow{2}{*}{$n=4$, $\tau=75\%$}} & AUC & 0.837 & 0.859 & 0.992 & 0.945  & \textbf{0.997}\\
\multicolumn{1}{c|}{} & $F_1$ & 0.686 & 0.684 & 0.941 & 0.960 & \textbf{0.973}\\
\noalign{\smallskip}
\hline
\noalign{\smallskip}
\multicolumn{1}{c|}{\multirow{2}{*}{$n=7$, $\tau=85\%$}} & AUC & 0.822 & 0.804 & 0.991 & 0.919  & \textbf{0.993}\\
\multicolumn{1}{c|}{} & $F_1$ & 0.528 & 0.511 & 0.897 & \textbf{0.941} & \textbf{0.941}\\
\noalign{\smallskip}
\hline
\end{tabular*}}
\end{table*}
\raggedbottom

\begin{table}[t]
\centering
\caption{Mean AUC values (in percentage) on CIFAR-10 using protocol 2.}
\label{table:cifar}
{\small
\begin{tabular}{ccccc}
\hline\noalign{\smallskip}
{Metric} & {OCSVM} & {OCGAN} & {LSA} & {ARAE}\\
\hline
\noalign{\smallskip}
AUC & 67.8 & \textbf{73.3} & 73.1 & 71.7 \\
\hline
\end{tabular}}
\end{table}
\raggedbottom

We present our AUC results for MNIST and Fashion-MNIST in Table \ref{table:mnist}. The table contains AUC values for each class as the normal class, which were achieved using protocol 2. Moreover, we report our results on the COIL-100 dataset in Table \ref{table:coil}. This table contains AUC and $F_1$ values for $n\in\{1,4,7\}$, where $n$ is the number of normal classes. We use protocol 1 for this dataset. For each $n\in\{1,4,7\}$, the percentage of the normal samples in the testing data ($\tau$) is defined in the table. The $F_1$ score is reported for the threshold value that is maximizing it.
As shown in Tables \ref{table:mnist} and \ref{table:coil}, we achieve state-of-the-art results in all of these datasets while using a  simpler architecture compared to other state-of-the-art methods, such as OCGAN, LSA, and GPND. Moreover, the results in Table \ref{table:coil} indicate that our method performs well when having multiple classes as normal. It also shows the low effect of the number of normal classes on our method performance. 

We also report our mean AUC results for the CIFAR-10 dataset using protocol 2, excluding the classes with AUC near 0.5 or below, in Table \ref{table:cifar}. Consider a classifier that labels each input as normal with probability $p$. By varying $p$ between 0 and 1, we can plot a ROC curve and compute its AUC. We observe that this method achieves an AUC of 0.5. So improvements below or near 0.5 aren't valuable (see \cite{zhu2010sensitivity} for more details). Consequently, classes 1, 3, 5, 7, and 9 which contained AUC values below 0.6 were excluded. As shown in the table, we get competitive results compared to other state-of-the-art approaches.

The AUC values of our method on the medical datasets are reported in Table \ref{table:medical}. We used $90\%$ of the normal data for training and the rest in addition to the anomalous data were used to form the testing data. Our method clearly outperforms other state-of-the-art approaches, which shows the effectiveness of our method on medical real-world tasks, where the dataset might be small and complex.

To show the stability of our training procedure, we compute the standard deviation of AUCs for the last 50 epochs of training. These values are reported for our method on MNIST in Table \ref{table:mnist} and for all the methods on Brain MRI - Tumor in Table \ref{table:medical}. From these tables, one can see the high stability of our training procedure. Moreover, It is apparent that our method is much more stable than other methods on the Brain MRI - Tumor dataset.

\begin{figure}[t] 
  \centering
  \includegraphics[width=0.85\linewidth]{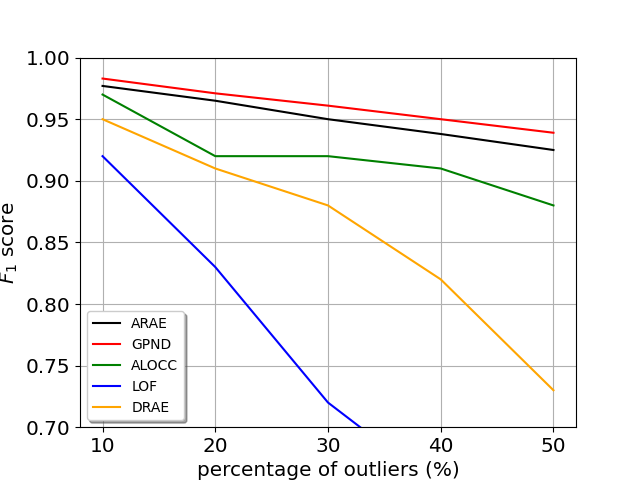}
  \caption{$F_1$ scores on the MNIST dataset using protocol 1, by taking the normal class as the positive one. }
  \label{fig:f1}
\end{figure}

We also evaluate our method using the $F_1$ score on the MNIST dataset. In this experiment, the normal class is the positive one. We use protocol 1 and vary $\tau$ between $50\%$ and $90\%$. We use $20\%$ of the training samples and sample from the anomalous classes to form a validation set with the same normal samples percentage as the testing data. This validation set is used to find the threshold that maximizes the $F_1$ score.
As shown in Figure \ref{fig:f1}, we achieve slightly lower $F_1$ scores compared to that of GPND. However, this figure shows the low impact of the percentage of anomalous data on our method performance.

\begin{figure}[t] 
  \centering
  \includegraphics[width=0.85\linewidth]{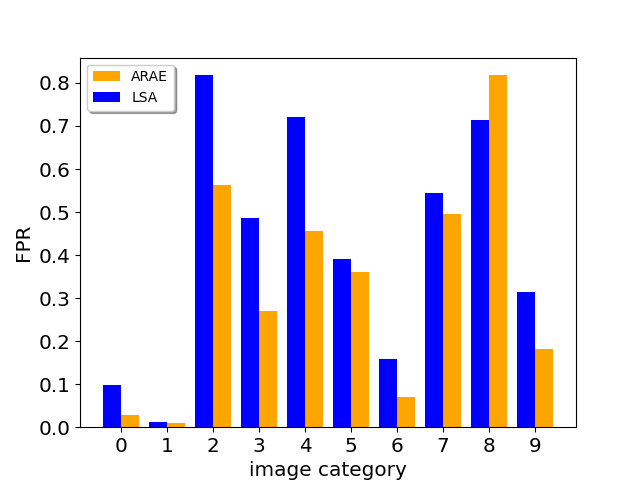}
  \caption{FPR at $99.5\%$ TPR on the MNIST dataset using protocol 2.}
  \label{fig:fpr}
\end{figure}

Furthermore, FPR values at $99.5\%$ TPR on the MNIST dataset using protocol 2, for \ac{arae} and LSA are compared in Figure \ref{fig:fpr}. One can see that despite having equal AUCs, \ac{arae} has lower FPR values compared to LSA and that it can reduce the FPR value more than $50\%$ in some cases.





\subsubsection{Adversarial Robustness}

To show the robustness of our model against adversarial attacks, we use PGD \cite{madry2017towards} with the $\epsilon$ parameter set to $0.05$ and $0.1$ on the reconstruction loss, to craft adversarial samples from the normal samples of the testing data. The normal samples of the testing data are replaced by the adversarial ones. The AUC results for this testing data are reported in Table \ref{table:adv} on the class 8 of the MNIST dataset, using protocol 2. As shown in the table, our method is significantly more robust against adversarial samples compared to LSA.

\subsection{Ablation}

\begin{table}[t]
\centering
\caption{AUC values for the attacked models. The values are reported for class 8 of MNIST using protocol 2.}
\label{table:adv}
{\small
\begin{tabular}{ccc}
\hline\noalign{\smallskip}
{Parameters} & {LSA} & {ARAE}\\
\hline
\noalign{\smallskip}
$\epsilon=0.05$ &  0.56 & \textbf{0.86} \\
$\epsilon=0.1$ & 0.17 & \textbf{0.76} \\
\hline
\end{tabular}}
\end{table}
\raggedbottom

\begin{table*}[t]
\centering
\caption{AUC values (in percentage) on MNIST using protocol 2. The results are reported for both one class and two classes as the normal data. Results for other variants of our method are reported.}
\label{table:union}
{\small
\begin{tabular*}{\textwidth}{c @{\extracolsep{\fill}} c @{\extracolsep{\fill}} c @{\extracolsep{\fill}} c @{\extracolsep{\fill}} c @{\extracolsep{\fill}} c @{\extracolsep{\fill}} c @{\extracolsep{\fill}} c @{\extracolsep{\fill}} c @{\extracolsep{\fill}} c @{\extracolsep{\fill}} c @{\extracolsep{\fill}} c}
\hline\noalign{\smallskip}
Method & 0 & 1 & 2 & 3 & 4 & 5 & 6 & 7 & 8 & 9 & Mean\\
\hline
\noalign{\smallskip}
DAE & 99.6 & 99.9 & 93.9 & 93.5 & 96.4 & 94.3 & 99.0 & 95.8 & 89.1 & 97.5 & 95.9\\
ARAE & 99.8 & 99.9 & 96.0 & 97.2 & 97.0 &
97.4 & 99.5 & 96.9 & 92.4 & 98.5 & 97.5\\
ARAE-A & 99.1 & 99.7 & 95.2 & 96.7 & 97.7 & 98.3 & 99.2 & 97.1 & 95.6 & 96.8 & 97.5\\
ARAE-R & 99.3 & 99.9 & 93.2 & 92.5 & 96.2 & 96.6 & 99.3 & 97.3 & 91.2 & 98.2 & 96.4\\
\hline
\hline\noalign{\smallskip}
Method & (4, 5) & (0, 7) & (1, 3) & (2, 6) & (8, 9) & (2, 9) & (0, 8) & (0, 1) & (2, 3) & (4, 9) & Mean\\
\hline
\noalign{\smallskip}
DAE & 88.8 & 94.1 & 98.2 & 90.3 & 86.8 & 91.8 & 91.1 & 99.7 & 90.0 & 97.3 & 92.8 \\
ARAE & 91.7 & 96.0 & 99.1 & 94.7 & 91.4 & 94.5 & 93.1 & 99.7 & 91.2 & 97.3 & 94.9 \\
ARAE-A & 95.0 & 97.1 & 97.4 & 95.7 & 91.5 & 92.6 & 94.3 & 98.8 & 94.3 & 97.4 & 95.4\\
\hline
\end{tabular*}}
\end{table*}

We train a \ac{dae}, as a baseline method, with a random uniform noise between 0 and $0.1$ using the same network as the one that is used in our approach.

Furthermore, In addition to the $\ell_\infty$ perturbation set, we consider $\ell_2$, and also rotation and translation perturbation sets. We need to solve a similar optimization to the one in Eq. \ref{summ_eq}, with the only difference being the perturbation sets \cite{engstrom2017exploring}. Specifically, we solve this optimization problem on $\ell_2$-bounded perturbations for each sample $X \in S$ through \ac{pgd} \cite{madry2017towards} again. We next solve this optimization on rotation and translation perturbation sets for each sample $X \in S$ by quantizing the parameter space, and performing a grid search on the quantized space and choosing the one with the highest latent loss. This is the most reliable approach for solving rotation and translation perturbations that is mentioned in \cite{engstrom2017exploring}. Following the approach in \cite{tramer2019adversarial}, we use the union of these perturbation sets to make the attack even stronger to avoid as much as brittle features that model might use \cite{ilyas2019adversarial}. We present our results on MNIST using protocol 2, in Table \ref{table:union}. This variant of our method is denoted as ARAE-A. Notably, the AUC is improved further in this variant in the most challenging class $8$ in MNIST from $92.4$ based on $\ell_\infty$ attack to $95.6$ using the union of the mentioned attacks. Despite this improvement, the average AUC is still the same as in the original ARAE method.

Instead of designing the attack based on the latent layer, one could directly use the reconstruction loss to do so. We denote this variant as ARAE-R. However, we observed that a model that is robust to the latter attack yields a lower improvement compared to ARAE (see Table \ref{table:union}). To justify this effect, we note that an \ac{ae} model that is robust based on the latter attack does not necessarily have a stable latent layer. This stems from the fact that the encoder and decoder are almost inverse functions by construction, and a destabilization of the latent encoding by an attack could be repressed by the decoder. In summary, an attack based on the latent layer is stronger than an attack based on the reconstruction error, and hence the former promotes more robust features.

We also report AUC values on MNIST by taking pairs of classes as the normal ones, in Table \ref{table:union}. These values show the improvement yield by both of the \ac{arae} variants. Note that when having multiple classes as normal, one should tune the $\epsilon$ parameter based on diversity and complexity of the training data.


\section{Visualization}
In the experiments section, we showed that our method improves the \ac{ae} performance and surpasses other state-of-the-art methods. In order to demonstrate the reasons behind this improvement, we show that \ac{arae} learns more semantically meaningful features than \ac{dae} by interpreting these two approaches.

\begin{table*}[t]
\centering
{\renewcommand{\arraystretch}{0.2} \small
    \begin{tabular*}{\textwidth}{c @{\extracolsep{\fill}} c @{\extracolsep{\fill}} c @{\extracolsep{\fill}} c @{\extracolsep{\fill}} c | c @{\extracolsep{\fill}} c @{\extracolsep{\fill}} c @{\extracolsep{\fill}} c @{\extracolsep{\fill}} c}
    \multicolumn{5}{c}{MNIST} & \multicolumn{5}{c}{Fashion-MNIST} \\
    \noalign{\smallskip}
    \hline
    \noalign{\smallskip}
    Input & \ac{dae} rec. & \ac{dae} map & \ac{arae} rec. & \ac{arae} map & Input & \ac{dae} rec. & \ac{dae} map & \ac{arae} rec. & \ac{arae} map \\
    \noalign{\smallskip}
    \hline
    \noalign{\smallskip}
    
    \includegraphics[width = 1.7cm , height = 1.7cm]{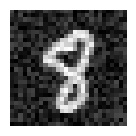} & \includegraphics[width = 1.7cm , height = 1.7cm]{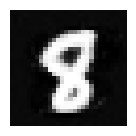} & \includegraphics[width = 1.7cm , height = 1.7cm]{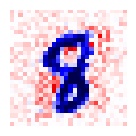} &
    \includegraphics[width = 1.7cm , height = 1.7cm]{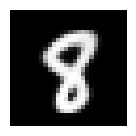} &
    \includegraphics[width = 1.7cm , height = 1.7cm]{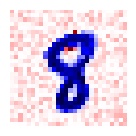} & \includegraphics[width = 1.7cm , height = 1.7cm]{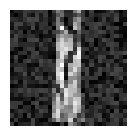} & \includegraphics[width = 1.7cm , height = 1.7cm]{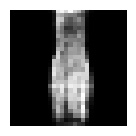} & \includegraphics[width = 1.7cm , height = 1.7cm]{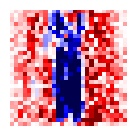} &
    \includegraphics[width = 1.7cm , height = 1.7cm]{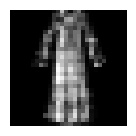} &
    \includegraphics[width = 1.7cm , height = 1.7cm]{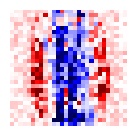} \\
    \includegraphics[width = 1.7cm , height = 1.7cm]{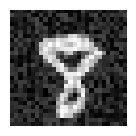} & 
    \includegraphics[width = 1.7cm , height = 1.7cm]{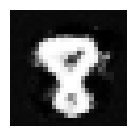} &
    \includegraphics[width = 1.7cm , height = 1.7cm]{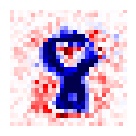} &
    \includegraphics[width = 1.7cm , height = 1.7cm]{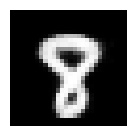} & 
    \includegraphics[width = 1.7cm , height = 1.7cm]{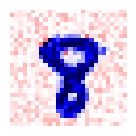} & \includegraphics[width = 1.7cm , height = 1.7cm]{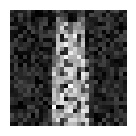} & 
    \includegraphics[width = 1.7cm , height = 1.7cm]{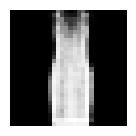} &
    \includegraphics[width = 1.7cm , height = 1.7cm]{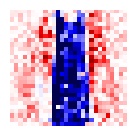} &
    \includegraphics[width = 1.7cm , height = 1.7cm]{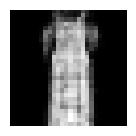} & 
    \includegraphics[width = 1.7cm , height = 1.7cm]{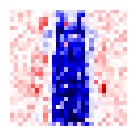} \\
    \includegraphics[width = 1.7cm , height = 1.7cm]{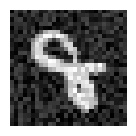} & 
    \includegraphics[width = 1.7cm , height = 1.7cm]{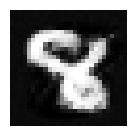} &
    \includegraphics[width = 1.7cm , height = 1.7cm]{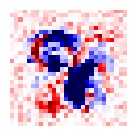} &
    \includegraphics[width = 1.7cm , height = 1.7cm]{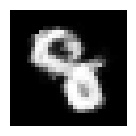} & 
    \includegraphics[width = 1.7cm , height = 1.7cm]{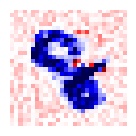} & \includegraphics[width = 1.7cm , height = 1.7cm]{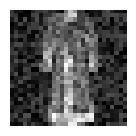} & 
    \includegraphics[width = 1.7cm , height = 1.7cm]{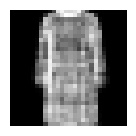} &
    \includegraphics[width = 1.7cm , height = 1.7cm]{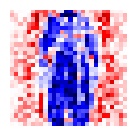} &
    \includegraphics[width = 1.7cm , height = 1.7cm]{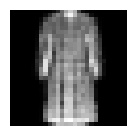} & 
    \includegraphics[width = 1.7cm , height = 1.7cm]{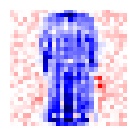} \\
    \includegraphics[width = 1.7cm , height = 1.7cm]{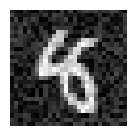} &
    \includegraphics[width = 1.7cm , height = 1.7cm]{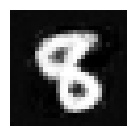} &
    \includegraphics[width = 1.7cm , height = 1.7cm]{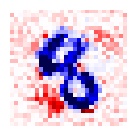} &
    \includegraphics[width = 1.7cm , height = 1.7cm]{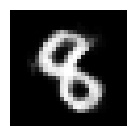} &
    \includegraphics[width = 1.7cm , height = 1.7cm]{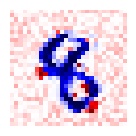} & \includegraphics[width = 1.7cm , height = 1.7cm]{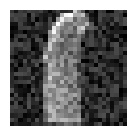} &
    \includegraphics[width = 1.7cm , height = 1.7cm]{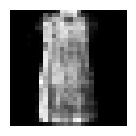} &
    \includegraphics[width = 1.7cm , height = 1.7cm]{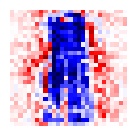} &
    \includegraphics[width = 1.7cm , height = 1.7cm]{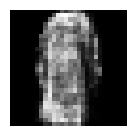} &
    \includegraphics[width = 1.7cm , height = 1.7cm]{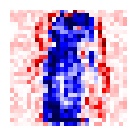} \\
    
    \includegraphics[width = 1.7cm , height = 1.7cm]{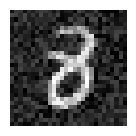} & 
    \includegraphics[width = 1.7cm , height = 1.7cm]{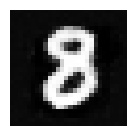} &
    \includegraphics[width = 1.7cm , height = 1.7cm]{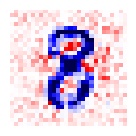} &
    \includegraphics[width = 1.7cm , height = 1.7cm]{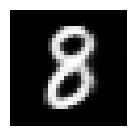} & 
    \includegraphics[width = 1.7cm , height = 1.7cm]{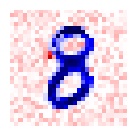} & \includegraphics[width = 1.7cm , height = 1.7cm]{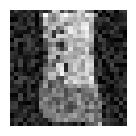} & 
    \includegraphics[width = 1.7cm , height = 1.7cm]{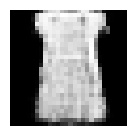} &
    \includegraphics[width = 1.7cm , height = 1.7cm]{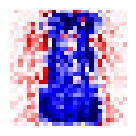} &
    \includegraphics[width = 1.7cm , height = 1.7cm]{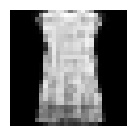} & 
    \includegraphics[width = 1.7cm , height = 1.7cm]{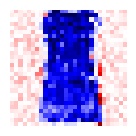} \\

    \end{tabular*}
}
    \captionof{figure}{\ac{arae} and \ac{dae} reconstructions and saliency maps for ten random inputs from MNIST and Fashion-MNIST datasets.}
    \label{map1}

\end{table*}
\raggedbottom

\begin{table*}[t]
\centering
{\renewcommand{\arraystretch}{1} \small
    \begin{tabular*}{\textwidth}{c | c @{\extracolsep{\fill}} c @{\extracolsep{\fill}} c @{\extracolsep{\fill}} c @{\extracolsep{\fill}} c @{\extracolsep{\fill}} c @{\extracolsep{\fill}} c @{\extracolsep{\fill}} c}
    \ac{arae} &
    \includegraphics[align=c, width = 1.7cm , height = 1.7cm]{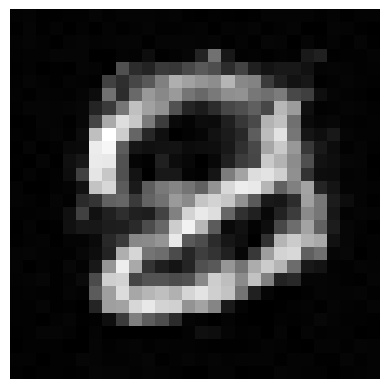} & 
    \includegraphics[align=c, width = 1.7cm , height = 1.7cm]{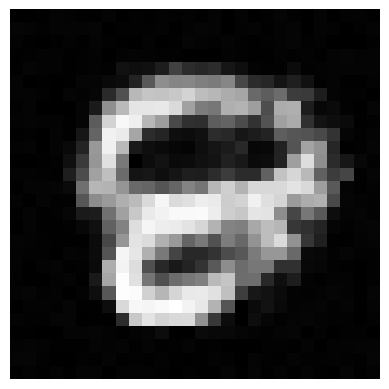} &
    \includegraphics[align=c, width = 1.7cm , height = 1.7cm]{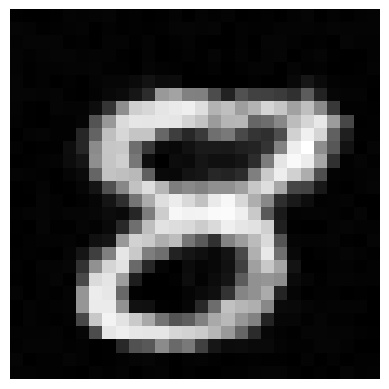} &
    \includegraphics[align=c, width = 1.7cm , height = 1.7cm]{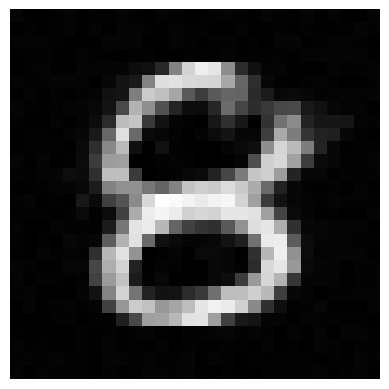} &
    \includegraphics[align=c, width = 1.7cm , height = 1.7cm]{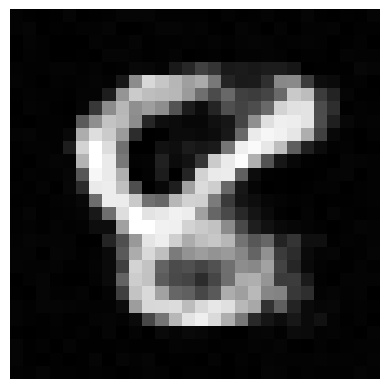} &
    \includegraphics[align=c, width = 1.7cm , height = 1.7cm]{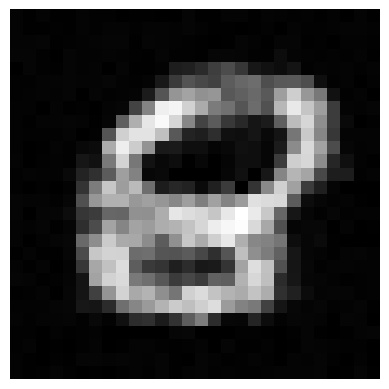} &
    \includegraphics[align=c, width = 1.7cm , height = 1.7cm]{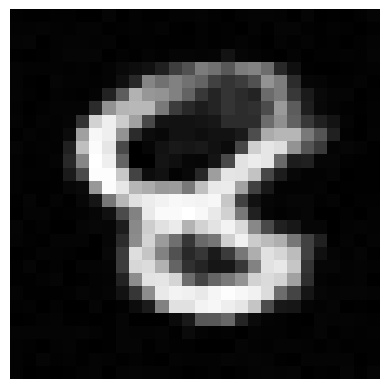} &
    \includegraphics[align=c, width = 1.7cm , height = 1.7cm]{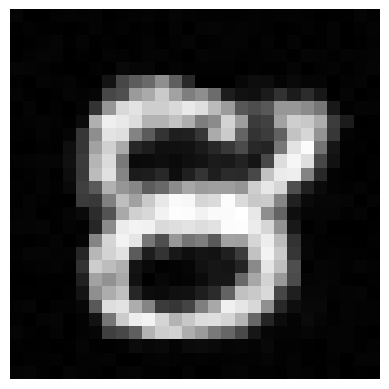} \\
    \noalign{\smallskip}
    \hline
    \noalign{\smallskip}
    \ac{dae} &
    \includegraphics[align=c, width = 1.7cm , height = 1.7cm]{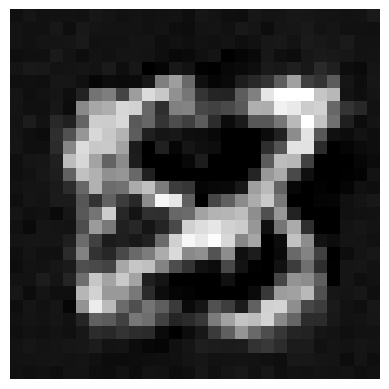} & 
    \includegraphics[align=c, width = 1.7cm , height = 1.7cm]{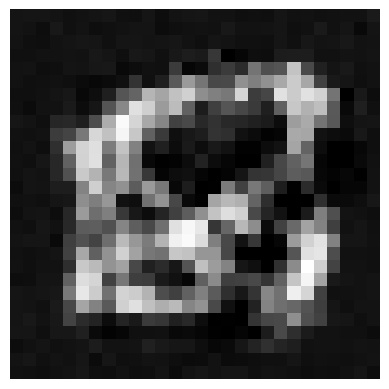} &
    \includegraphics[align=c, width = 1.7cm , height = 1.7cm]{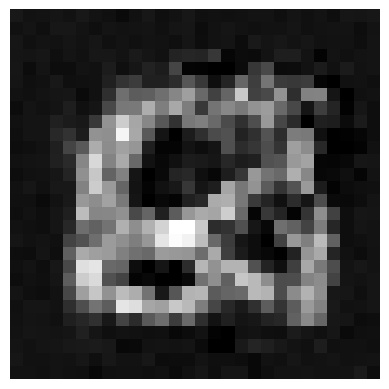} & 
    \includegraphics[align=c, width = 1.7cm , height = 1.7cm]{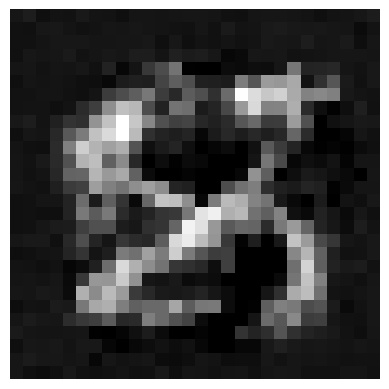} &
    \includegraphics[align=c, width = 1.7cm , height = 1.7cm]{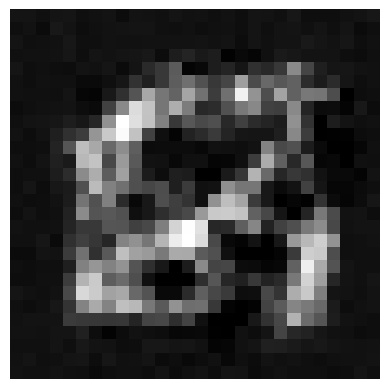} & 
    \includegraphics[align=c, width = 1.7cm , height = 1.7cm]{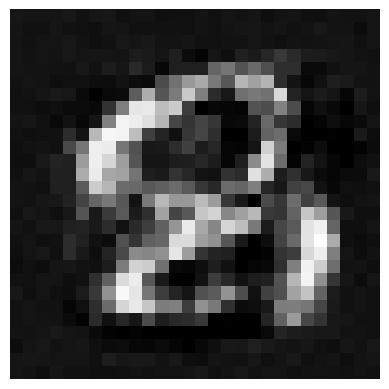} &
    \includegraphics[align=c, width = 1.7cm , height = 1.7cm]{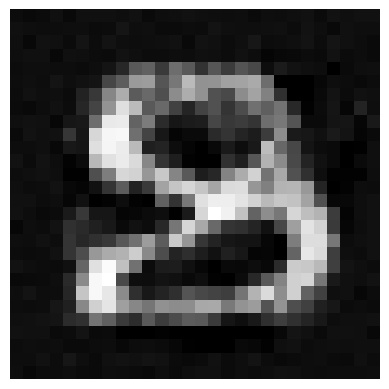} & 
    \includegraphics[align=c, width = 1.7cm , height = 1.7cm]{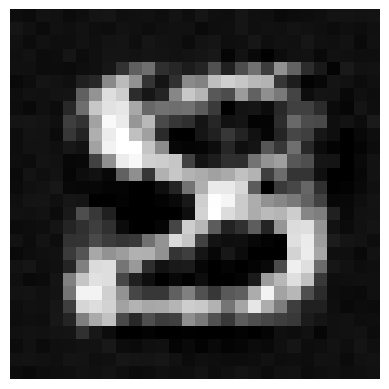} \\
    
    \end{tabular*}
}
    \captionof{figure}{Local minima of inputs of \ac{arae} and \ac{dae}, by initializing the input with random noise and optimizing the reconstruction loss with respect to the input. \ac{arae} produces more realistic $8$ digits compared to \ac{dae}.}
    \label{local1}

\end{table*}
\raggedbottom

\subsection{Interpreting with Occlusion-1}
In this method, we measure the effect of each part of the input on the output, by occluding it and observing the difference in the output. Finally, we visualize these differences as a saliency map \cite{zeiler2014visualizing, ancona2016towards}. In the occlusion-1 method, we iteratively set each pixel to black and then observe the reconstruction error. If it increases, we set the corresponding pixel in the saliency map to blue, and otherwise, we set it to red. The intensity of a pixel is determined by the amount that the reconstruction error has changed. We compare \ac{arae} and \ac{dae} reconstructions and saliency maps on MNIST and Fashion-MNIST datasets, in Figure \ref{map1}.

For the MNIST dataset, the model has been trained on the class $8$ and noisy inputs are obtained by adding a uniform noise in the interval $[0,0.4]$. The outputs and saliency maps of \ac{arae} and \ac{dae} are shown for five random inputs in the normal class. It is evident that \ac{dae} is focusing too much on the random noises and has a poorer reconstruction than our model.

Similar to MNIST, we carry out the occlusion-1 method on the class dress of the Fashion-MNIST dataset. For Fashion-MNIST, it is also obvious that random noises have a larger effect on the output of \ac{dae}. Furthermore, \ac{dae} reconstructions are less accurate than those of \ac{arae}. These observations are consistent with the known fact that adversarial robustness can increase the model interpretability \cite{tsipras2018robustness} by avoiding the learning of brittle features \cite{ilyas2019adversarial}.

    

\subsection{Local Minima Visualization}

We expect from an ideal model that is trained on the MNIST class $8$, to have a lower reconstruction error as the input gets more similar to a typical $8$. With this motivation, we start from random noise and iteratively modify it in order to minimize the reconstruction error using gradient descent. The results achieved by our model and \ac{dae} are shown in Figure \ref{local1}. This figure demonstrates that inputs that lead to local minima in \ac{arae} are much more similar to $8$, compared to \ac{dae}.

\section{Conclusions}

We introduced a variant of \ac{ae} based on the robust adversarial training for novelty detection. This is motivated by the goal of learning representations of the input that are almost robust to small irrelevant adversarial changes in the input. A series of novelty detection experiments were performed to evaluate the proposed \ac{ae}. Our experimental results of the proposed \ac{arae} model show state-of-the-art performance on four publicly available benchmark datasets and two real-world medical datasets. This suggests that the benefits of adversarial robustness indeed go beyond security. Furthermore, by performing an ablation study, we discussed the effect of multiple perturbation sets on the model. Future work inspired by this observation could investigate the effect of other types of adversarial attacks in the proposed framework.

\end{document}